\documentclass[letterpaper]{article} 
\usepackage{DeLF_CR}  
\usepackage{times}  
\usepackage{helvet}  
\usepackage{courier}  
\usepackage[hyphens]{url}  
\usepackage{graphicx} 
\usepackage{natbib}  
\usepackage{caption} 
\usepackage{algorithm}
\usepackage{algorithmic}
\usepackage{algorithm}
\usepackage{algorithmic}
\usepackage{amsmath, amssymb}
\usepackage{pythonhighlight, listings}
\usepackage{xcolor}
\usepackage{comment}

\definecolor{codegreen}{rgb}{0,0.6,0}
\definecolor{codegray}{rgb}{0.5,0.5,0.5}
\definecolor{codepurple}{rgb}{0.58,0,0.82}
\definecolor{backcolour}{rgb}{0.95,0.95,0.92}

\lstdefinestyle{mystyle}{
    backgroundcolor=\color{backcolour},   
    commentstyle=\color{codegreen},
    keywordstyle=\color{magenta},
    numberstyle=\tiny\color{codegray},
    stringstyle=\color{codepurple},
    basicstyle=\ttfamily\footnotesize,
    breakatwhitespace=false,         
    breaklines=true,                 
    captionpos=b,                    
    keepspaces=true,                 
    numbers=left,                    
    numbersep=5pt,                  
    showspaces=false,                
    showstringspaces=false,
    showtabs=false,                  
    tabsize=2
}

\usepackage{newfloat}
\usepackage{listings}
\DeclareCaptionStyle{ruled}{labelfont=normalfont,labelsep=colon,strut=off} 
\floatstyle{ruled}
\newfloat{listing}{tb}{lst}{}
\floatname{listing}{Listing}
\title{DeLF: Designing Learning Environments with Foundation Models}
\author{
    Aida Afshar and Wenchao Li
}
\affiliations{
    Boston University\\

    \{aafshar,wenchao\}@bu.edu
}

\usepackage{bibentry}

\begin{document}

\maketitle

\begin{abstract}
Reinforcement learning (RL) offers a capable and intuitive structure for the fundamental sequential decision-making problem. Despite impressive breakthroughs, it can still be difficult to employ RL in practice in many simple applications. In this paper, we try to address this issue by introducing a method for designing the components of the RL environment for a given, user-intended application. We provide an initial formalization for the problem of RL component design, that concentrates on designing a good representation for observation and action space. We propose a method named DeLF: Designing Learning Environments with Foundation Models, that employs large language models to design and codify the user's intended learning scenario. By testing our method on four different learning environments, we demonstrate that DeLF can obtain executable environment codes for the corresponding RL problems. 

\end{abstract}

\begin{figure*}[ht!]
\centering
\includegraphics[width=\textwidth]{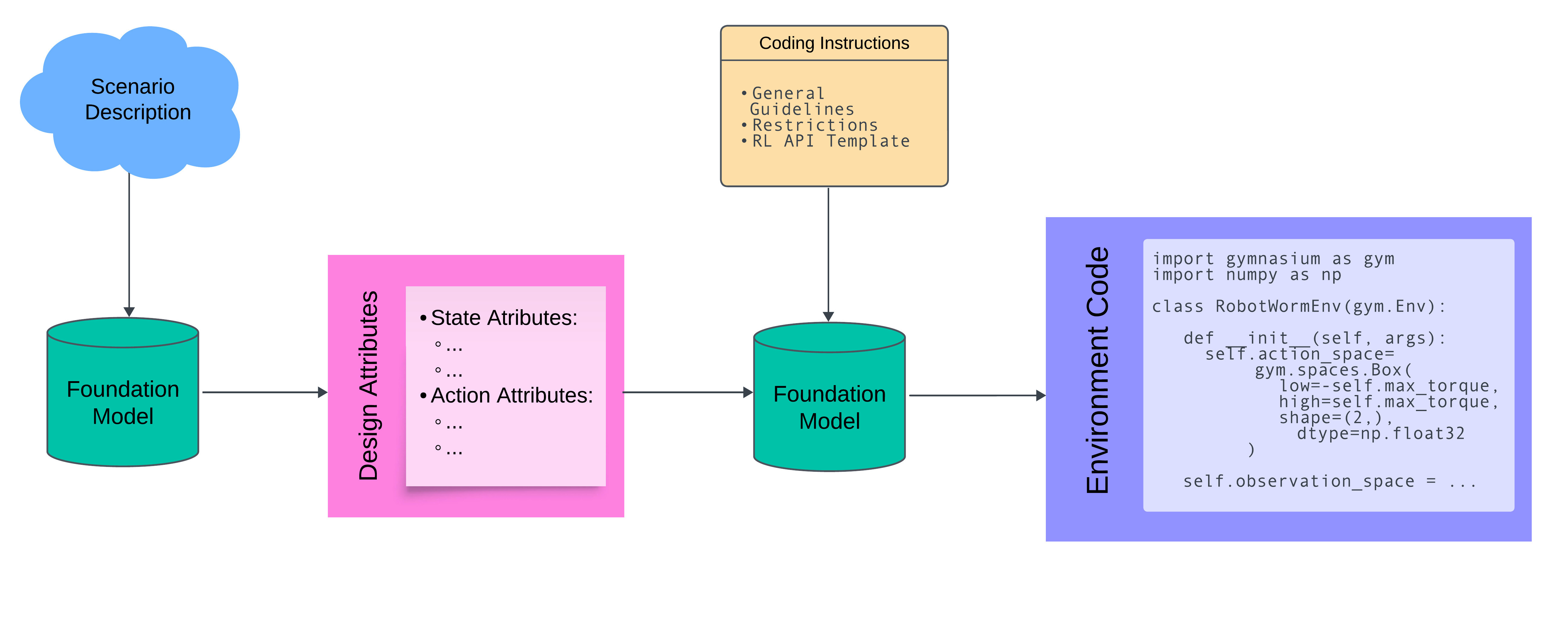} 
\caption{Environment design with DeLF: The user provides a description of a learning scenario to the foundation model (e.g. large language models); the foundation model proposes a design for observation and action attributes; By having the basic template of the user's desirable RL API as context, DeLF is able to generate an initial sketch of the environment code that can be fed into an RL Algorithm.}
\label{fig1}
\end{figure*}

\section{Introduction}
Reinforcement learning is a powerful paradigm for training intelligent agents through interactions with their environment. 
Most recent efforts have been dedicated to improving different aspects of RL, such as sample efficiency, reward design, and partial observability. Various environments have also been introduced to showcase the exciting potential of RL. These environments span a spectrum of non-realistic toy examples to realistic safety-critical simulations \cite{6313077, bellemare2013arcade, tai2023pyflyt}. Ultimately, RL or any other decision-making framework is going to be applied to a learning scenario where all these theoretical and experimental efforts come into practice. Therefore, it's critical to investigate how we design and codify various components of the RL environment for a given user-intended application. 

Most of the time, researchers and developers go through a repetitive cycle of trial-and-error, changing representations of observation and action space to finally see some indications of learning. Throughout employing RL as a method to learn a specific task, we might need to figure out a reach-enough representation of the observation space that gives adequate information to the agent for learning the task. Additionally, the agent might be able to do complex actions and perform a diverse range of motions. For example, the action space of a human-like robotic arm is usually a vector of the continuous values of the torque of each joint, leading to a high-dimensional action space. While in practice, most of the state-of-the-art RL algorithms are unable to handle such large action spaces. This raises the question of what is a good representation of the observation and action space in a reinforcement learning problem.  As explained, we have a variety of design choices for codifying the environment, but mostly no guaranteed path to find the right design choice. The exhaustive cycle of trial and error reduces the hope of conveniently employing RL in diverse applications. This applies not only to researchers and developers but more importantly to users with less coding and technical skills; considering that experts already have a good intuition on how to define RL components such as observation, action, and reward. Hence, RL can vastly benefit from the tools that facilitate the work prior to training, such as the automated implementation of the learning scenario in a structure that is executable by the RL algorithm.

The notion of representation and the problems surrounding defining a good representation goes well beyond RL and the sequential decision-making domain and the term can be studied for the mathematical modeling of any dynamical system. A state space representation characterizes the system in the mathematical language and portrays its evolution through a timespan. It is the representation that bridges reality to theory and enables us to study the system in mathematical language; hence, it's important to study different characteristics of a representation and provide theoretical guarantees in addition to practical tools that help us design better representations for our methodologies.

In this paper, we focus on designing observation and action representations for a given RL problem. We specify some properties for the \emph{goodness} of a design choice for the observation and action representation. We utilize foundation models as an assistant to help us design and extract better representations for a sequential decision-making scenario that is going to be learned by RL. Finally, we propose DeLF: \textbf{De}signing \textbf{L}earning Environments with \textbf{F}oundation Models, a method that takes the first steps toward using this concept in practice. DeLF mainly concentrates on extracting a sufficiently good observation and action representation from the task description. After generating and evaluating the design choice of these representations, we can have an initial sketch of the gym-like environment that can be fed into an RL algorithm.  Foundation models can be well-suited assistants to address this gap, as they are inherently designed to compute a representation of the inputs of various modalities. In this paper, we utilized large language models since their in-context learning capabilities together with rich enough prompts can give us a pool of options to design the representation of different components in an RL problem. In addition to that, there is a convenient flow from language description to code in large language models specialized for code generation such as GPT-4 that helps us to generate a sketch of an RL environment code. We summarize our contribution as follows.
\begin{enumerate}
    \item Providing a formalization for the problem of RL Component Design, 
    \item Specifying the properties of sufficiency and necessity for observation and action representation in RL; 
    \item Introducing the notion of \emph{component extraction function}, an operator for extracting the design choice of various components in a learning paradigm.
    \item Introducing DeLF, a method for designing and codifying RL components with large language models.
\end{enumerate}

Despite the fact that DeLF doesn't directly focus on reward design, we envisage that it can be used in synergy with the recent successful results on designing reward functions with language models \cite{ma2023eureka}. This can fulfil the possibility to generate the RL environment code for an arbitrary learning environment; without the user having major coding skills and instead interacting with the component extraction function, where in this paper is a large language model. Finally, We test our method DeLF on some of the well-known RL scenarios. We open source all the codes, prompts, and experiment results for future studies in https://github.com/AidaAfshar/DeLF.

\section{Prelimineries}

\subsection{Foundation Models} 
The term foundation model often refers to an embedding function trained on a massive dataset, capable of performing various tasks. These embedding functions usually model a conditional probability on a large dataset. In practice, the dataset can be a sequence of text tokens, images, audio, or a combination of these modalities.

Foundation models were initially popularized by transformer-based language models which factorize a joint distribution over a sequence of the input format auto-regressively. The core capability of transformers comes from the self-attention mechanism; a mechanism that computes a representation of an input sequence by making use of the order of the sequence via positional encoding \cite{brown2020language}.

\subsection{Reinforcement Learning}
Reinforcement learning is a model of the sequential decision-making problem, where an agent interacts with an environment to perform a specific task. This interaction usually happens in a sequence of discrete timesteps where in each timestep, an agent chooses an action, receives a reward, and transitions to a new state of the environment accordingly. In most of the realistic scenarios, the RL problem is formalized as a Partially Observable Markov Decision Process, POMDP which is a tuple $\langle S, A, P, R, O, \Omega \rangle$ where $S$ is the state space, $A$ is the action space, $P: S \times A \rightarrow \mathbb{P}[S]$ is the transition function, $R: S \times A \rightarrow \mathbb{R}$ is the reward function, and $O$ is the agent's observation space, and $\Omega: S \times A \rightarrow \mathbb{P}[O]$ is the observation function \cite{KAELBLING199899}. We call each member of this tuple, a \emph{component} of the RL Problem, and we try to find the proper design choice for the observation and action components with DeLF.

\section{Problem Setting}

\section{Definitions}
\textbf{Description Space} $D$ is a set of all possible descriptions for a learning environment. A description $d \sim D$ can be in the form of human-language text, an image or video of the agent performing the task in the environment, audio, etc.
\\
\textbf{Component Attributes} $Att_{C}$ is the set of attributes we need to describe the component $C$.
\begin{itemize}
\item \textbf{Observation Attributes} $Att_{O}$ is the set of all the attributes we need to describe what the agent is observing in the environment.
\item \textbf{Action Attributes} $Att_{A}$ is the set of all the attributes we need to describe the possible actions of the agent in the environment.
\end{itemize}
\textbf{Attribute Quantification}: For each attribute $a$, we define a quantification $Q_a$ which is a numerical representation of the attribute. This numerical representation is usually in two forms:
\begin{enumerate}
    \item Continuous Quantification
    \begin{align}
        Q \subseteq [l, u]^n \qquad l,u \in \mathbb{R}  \qquad \text{e.g.} \quad [-1,1]^n
    \end{align}
    \item Discrete Quantification
    \begin{align}
        Q \subseteq \mathbb{Z} \qquad\qquad \text{e.g.} \quad \{0,1\} 
    \end{align}
\end{enumerate}
\textbf{Design Choice of Component C} is the tuple $\langle Att_C, Q_C \rangle$ where $Att_C$ is the set of component attributes and $Q_C$ is the set of quantification assigned to each attribute in $Att_C$. 
\\
\textbf{Task}: Task $\tau$ is a description of what the agent is supposed to do in the environment. As mentioned before, a description can be in the form of human-language text, an image or a video of the agent performing the task, or a mathematical objective function. etc.
\\
We introduce four notations $\vdash, \nvdash, \Vdash, \nVdash$ for task $\tau$. 
We write, 
\begin{align}
    \langle c_1, c_2, ..., c_n \rangle \vdash \tau
\end{align}
if \emph{the design choices of components $ \langle c_1, c_2, ..., c_n \rangle$ leads to a successful learning of the task $\tau$.} We write, 
\begin{align}
    \langle c_1, c_2, ..., c_n \rangle \nvdash \tau
\end{align}
if this design choice does not lead to a successful learning of the task. We write, 
\begin{align}
    c \Vdash \tau 
\end{align}
if given the proper design choices of all other components, the design choice of component $c$ leads to successful learning of the task $\tau$. We write, 
\begin{align}
    c \nVdash \tau
\end{align}
if this choice does not lead to a successful learning of the task.
\\
By the term successful learning, we mean that the agent is acting according to a (near) optimal policy. In practice, a successful learning is usually determined by analyzing the performance metrics and the agent's behavior on the learned policy. 
\\
\textbf{Sufficient Observation Space}: The representation of observation space $O$ is called \emph{sufficient with respect to task $\tau$} if given all other components designed properly, $O$ leads to a successful learning of the task $\tau$.
\begin{align}
    \text{$O$ is sufficient} \iff O \Vdash \tau
\end{align}
\textbf{Sufficient Action Space}: The representation of action space $A$ is called \emph{sufficient with respect to task $\tau$} if given all other components designed properly, $A$ leads to the successful learning of the task $\tau$.
\begin{align}
    \text{$A$ is sufficient} \iff A \Vdash \tau
\end{align}
\textbf{Necessary Observation Space}: The representation of observation space $O$ is called \emph{necessary with respect to task $\tau$} if it is the minimal subset of the observation space required for learning the task $\tau$;  
\begin{align}
    O \Vdash \tau \quad \text{and} \quad O \setminus \{v\} \, \nVdash \tau \qquad \forall \; v \in O 
\end{align}
\textbf{Necessary Action Space}: The representation of action space $A$ is called \emph{necessary with respect to task $\tau$} if it is the minimal subset of the action space required for learning the task $\tau$;  
\begin{align}
    A \Vdash \tau \quad \text{and} \quad A \setminus \{v\} \, \nVdash \tau \qquad \forall \; v \in A
\end{align}
\textbf{Component Extraction Function} $\hat{C}: D \rightarrow \langle Att_C, Q_C\rangle$ is a function that maps the description space to the space of design choices for the representation of the component $C$. We call this operator a \emph{component designer} who extracts a design choice for component $C$ out of description $d \sim D$.

\section{The Problem of RL Component Design}
We formalize the problem of designing a learning environment as a tuple $\langle \hat{O}, \hat{A}, \hat{R}, I \rangle$ where $\hat{O}: D \rightarrow O$ is the observation extraction function, $\hat{A}: D \rightarrow A$ is the action extraction function, $\hat{R}: D \rightarrow R$ is the reward extraction function e, and finally $I: S \times A \rightarrow S$ is the agent-environment interaction function. \footnote{The function $I$ majorly replicates the agent's transition in the environment and is often referred to as \emph{step} function in the environment code. The function $I$ mainly relies on the transition dynamics of the model, which is not the focus of this paper. We will assume that the agent can interact with the environment through simulation or real-world interaction.}

\section{Language Models as RL Component Designers}
Here, we use language models as the extraction function for the attributes, the observation space, and the action space. The input for this extraction function will be the human-language description of the task. The output will be the set of attributes, the codified definition of observation space, and the codified definition of action space, respectively. Practically, we need to provide a context where we explain the general guidelines and templates that we want the language model to follow while generating the outputs. These guidelines can vary based on the use-case and do not restrict the user to follow a certain context template. For clarification, we provide the original prompts that we used as context for our experiments in Appendix B.

\section{Method}
DeLF consists of three sections which we refer to as \textbf{ICE}; \textbf{I}nitiation, \textbf{C}ommunication, and \textbf{E}valuation. A sufficient design of an environment with a language model is reachable via ICE.
\subsection{DeLF Initiation}
To get the desired learning environment code with fewer prompts, we find it helpful to divide the initial query to the language model into two parts:
\begin{itemize}
\item Design: Describe the environment and task to the language model and ask it to extract the observation and action space.
\item Codify: Provide the code structure (the intended RL API we want to use to train the agent) as a context for the language model. Optionally, general coding guidelines that we expect the language model to follow.
\end{itemize}
We observed in our experiments that asking for the design choices of observation and action representation separate from the codify query will significantly improve the speed and convenience of using DeLF. This is possibly due to the reason that language models tend to get lost in the middle \cite{liu2023lost}, and providing all the description and coding details in one prompt will decrease the significance of the design choice for the language model.  This distinction will let the language model focus on a more accurate extraction of observation and action attributes first and then get involved in coding details.

\subsection{DeLF Communication}
It's Ideal to have a zero-shot method for generating the right representations that are sufficient and necessary for the agent to learn the task. Such methods are often equipped with an evaluation metric that checks these desirable metrics in the representation. To our knowledge, there is no systematic method to evaluate these metrics for a given representation, and RL practitioners design these representations often empirically and by relying on their domain-specific knowledge. Hence, for a method like DeLF, communication is key. The user can leverage their own domain knowledge or intuition of representations and correct any obvious mistake or hallucination of the language model after the Design stage. Also, the codified environment produced by DeLF might encounter programming errors and bugs in running time. The debugging effort of these errors can be uplifted and corrected by communicating them with the language model. The benefit of separating design and codification stages can be seen more clearly here as the errors faced at execution time are distinct from the design of state and action representation and are internal to implementation details. The communication step also helps drastically with aligning the user's intended scenario with the outcome of representation design choices. 

\subsection{DeLF Evaluation}
Evaluation is a critical step to assess the correctness and practicality of any method, including DeLF. As explained in the previous section, there is a lack of evaluation metrics that can assess some of the desirable properties of a component representation in RL literature. In the case of our problem, we ideally want to answer \emph{what} is a good observation or action representation for a given RL task, and \emph{how} should we measure this notion of goodness? To answer this question, we refer to the properties introduced in the Problem-Setting section. We want the state representation to have adequate information for the agent to learn the task successfully. Here we need to fix the interpretation of successful learning. We interpret the successful learning of the task as when the agent is behaving according to an optimal policy. This optimal policy is most often not accessible in practice and should be learned and approximated through various methods like RL which often require a large number of samples to attain the suboptimal policy. This makes it difficult to theoretically evaluate the representations prior to training efficiently. In this paper, we try to evaluate the practicality of using DeLF by two factors: i) the number of words that an individual user needed to explain in the DeLF initiation step. ii) Number of communication trials including the trial needed to improve the representation designs in addition to the debugging trials required to reach the executable environment code. Introducing a more mathematically solid metric and elaborating on the available is explained more in the Discussion and Future Work section.

\section{Experiments and Results}

\begin{table*}[t]
\centering
\begin{tabular}{|p{4cm}|p{4.5cm}|p{3cm}|p{3cm}|}
\hline
\multicolumn{4}{|c|}{Environment Design with DeLF} \\
\hline
\textbf{Environment} & Observation and Action Space & Description Tokens& Trials to Execution  \\
\hline
\textbf{Recommender System} & Hybrid & 104 & 3\\
\hline
\textbf{Self-Driving Car} & Hybrid & 135 & 6\\
\hline
\textbf{Swimmer} & Continuous & 98 & $<$10\\
\hline
\textbf{Key-Lock} & Discrete & 48 & 2\\
\hline
\end{tabular}
\caption{Results of DeLF experiments on four learning scenarios. The description token is different from the total number of tokens for each experiment by a fixed number. Trials to execution is the number of extra communication queries plus debugging steps needed to reach an executable environment code.}
\label{table1}

\end{table*}

In this section, we test DeLF for designing three environments with different observation and action characteristics. The example ICE prompts for one of the environments are available in Appendix B.

\subsection{Recommender System}
We described a simple recommender system that recommends several products to the user one by one, and the user will decide to either buy or pass the product. Each product is associated with a fixed number of features that explain the product the best. The recommender system is supposed to learn which features matter the most to the user and recommend products that are most likely for the user to buy.

One of the key attributes to include in the representation of the observation space in the recommender system application is the \emph{history} of the user's purchase, which is necessary for the agent to learn the task. Otherwise including the user's decision for the most recent purchase would be insufficient for RL. This technique is usually used to combat the limitations of Markovian property while applying RL to different applications. We observed that DeLF can correctly capture this attribute of observation space in the first shot. The were three communication and debugging trials in total; one of them was a misdesign of one attribute in the action representation and the two others were minor debugging problems.

\subsection{Self-Driving Car}
We designed a learning scenario for a self-driving agent in a 2-lane street. The agent can accelerate or brake but should stay below a certain speed limit. There are some obstacles placed in the street which the agent must avoid. The task is to reach a certain destination while avoiding obstacles and overspeeding. The closest expert-designed environment to this scenario is HighwayEnv \cite{highway-env}.

Despite providing a basic description of the scenario, GPT-4 produced a significantly relevant environment that was ready to execute after a few communication queries. All of the mistakes were considered minor except for one that violated one of the coding rules specified in the codify query.

\subsection{Swimmer}
The swimmer environment \cite{coulom2002reinforcement}, implemented and popularized as one of the MuJoCo environments \cite{todorov2012mujoco}, consists of three segments connected to each other by two joints. The agent is able to move by creating torque for each of these joints and the friction caused by the underlying surface. 

We use a different name in our description to intentionally reduce the reliance of GPT-4 on the MuJoCo Environment while generating the code. The user input was written based on the basic understanding of one of the authors of the environment. GPT-4 proposed a relatively accurate design choice for observation and action spaces, very close to the expert-designed version of the environment. Besides that, it produced the environment code with less than 10 debugging trials. 

\subsection{Key-Lock}
Grid World environment \cite{MinigridMiniworld23}, an $n \times n$ surface with an agent able to move one grid to any of the 4 main directions in each timestep. It's possible to define various scenarios in this environment; here we focus on the key-lock version, where there is a key and a lock placed in two different grids of the environment. The agent is supposed to first find the key, and then find and open the lock.

In our experiments, despite providing a relatively naive description of the problem, GPT-4 could generate an executable environment code in two trials. Both the action and observation attributes extracted by GPT-4 are compatible with the original design and our intuitive understanding of the problem. The two debugging trials were due to minor coding mistakes, such as argument mismatch.

\section{Discussion and Future Work}
\subsection{Different Modalities of Foundation Models}
The formalization of the extraction function provided before is not limited to language models. One can imagine that by the time we have capable foundation models for images or videos, we can use them as extraction functions in the problem of RL component design. For instance, it would be ideal to provide the video of the environment and the embodied agent in the input and ask the foundation model to extract the observation and action representations.

\subsection{Evaluation Metrics}
There is a lack of evaluation metrics in the literature that assess the design of a representation concerning the goal it wants to achieve. In the case of RL, we want the design choice of observation and action representation to be sufficient and necessary for learning the task as defined in the Problem Setting section. This can be hard and costly to investigate since finding the (sub)optimal policy or learning an approximation of it is a fundamentally difficult problem, even when we have access to the underlying model of the environment. Recently, \cite{laidlaw2023bridging} came up with a new metric named effective horizon that might give new insights into both the theoretical and practical assessment of state and action representation.

\subsection{Reward Design with Language Models}
DeLF together with the recent works on using language models for reward design \cite{yu2023language} can drastically ease the definition and implementation of major RL components. In this sense, DeLF is complementary to Eureka \cite{ma2023eureka} since the latter gets the environment code as input and generates the reward function for the environment. Hence, it's interesting to see how the synergies between these two might work in order to generate the right design choice for RL components based on user description.

\subsection{Language models specialized for producing gym-like environments}
In this paper, we tested our method on GPT4 with no specific fine-tuning or pertaining. The result of our experiments on four environments is further encouraging our initial claim that foundation models can be good component extraction functions by design. On the other hand, we showed that codifying these representations into a gym-like environment is achievable through large language models specialized for coding with some communication/debugging steps. These communication steps might increase with the complexity of the task description and the learning environment. We suspect that pretraining or finetuning these models on a relevant dataset can vastly improve the result. Hence, it would be useful to gather a dataset of previously implemented gym-like environments. This dataset can then be used to pre-train or fine-tune the language model and improve its ability on the specific task of codifying the gym-like environment.

\section{Conclusion}
In this paper, we took a different approach to studying the observation and action representation in RL. We first formalized the problem of RL component design by introducing the notion of component extraction function. Then, we discussed that foundation models can be powerful candidates for the extraction function, due to their abilities to process various user inputs and generate relevant representations of the input sequence. We tested this idea on large language models, by using GPT-4 as the extraction function for observation and action space. Ultimately, we introduced DeLF, a method for designing observation and action representation and codifying an initial sketch of the RL environments. DeLF showed successful results on four different learning scenarios, generating executable environment codes after a few communication and debugging trials. We tried to take a step forward toward a more practical and broad usage of RL in various applications, and we hope these results create motivation for possible extensions of this idea.

\bibliography{DeLF_CR.bib}

\newpage

\section{Appendix}
\subsection{A. Codes}
The following GitHub repository contains the GPT-generated codes: https://github.com/AidaAfshar/DeLF. We provide the code for one of the experiments in Appendix C as well. 

\subsection{B. Design and Codify queries}

Here are the original prompts we used in our experiments. We followed the same template as shown below for all of our experiments.

\textbf{B.1. Design Query}
\\
The design prompt is basically the description of the problem provided by the user, and queries the user to extract the design attributes. 
\\
\begin{lstlisting}
I want to train a Reinforcement Learning Agent for a learning scenario, and I want you to design the RL environment components of this scenario. 

Here is the scenario:
<Describe the scenario here> 

What's the observation space and action space for this scenario? List the attributes of observation and action and justify why these sets of attributes are enough for the agent to learn the task in this environment.
\end{lstlisting}    
\textbf{B.2. Codify Query}
\\
The codify query contains general coding instructions and the environment code format from OpenAI gym API. The code format is basically for enforcing the input-output format of major functions. This makes the environment code directly executable with commonly used RL algorithms.
\\
\begin{lstlisting}
I want you to codify the environment for the scenario I described before. I want you to follow the format of the OpenAI gymnasium environments library provided below. It's recommended to start with "import gymnasium as gym".
Fill out the functions and generate code where specified so that we can train this agent. You can add new functions if you want but Don't change input-output format of the existing functions. 
You should initialize self.observation_space and self.action_space in the init function. Use the predefined spaces from gymnasium.spaces to initialize them. 
You are free to use Python libraries such as numpy in the environment. Just remember to import them.



class EnvName(gym.Env):
  
    def __init__(self, args):
    	
        # Generate Code Here



    def _perform_transition(self, action):
        
        # Generate Code Here
        
        return  observation
	
    
    
    def _get_reward(self, action, observation, terminated):
    	
    	# Generate Code Here
    	
    	return r
    
    
    def _is_terminated(self):
    	
    	# Generate Code Here
    	
    	return terminated
    
    
	def _get_info(self):
		
		# Generate Code Here If Necessary
		
    	return info    
    

	def step(self, action):
    
    	observation = self._perform_transition(action)
    	terminated = self_is_terminated(observation)
    	reward = self._get_reward(action, observation, terminated)
    	info = self._get_info()

    	return observation, reward, terminated, False, info
    
    
    
    
    
    def reset(self, seed=None, options=None):
    	super().reset(seed=seed)

		# Generate Code Here
	
    	return observation, info
\end{lstlisting}

\subsection{C. Experiments}

\textbf{C.1. GPT-generated code for the Recommender System Environment}
\\
\begin{lstlisting}[language=python]
import gym
from gym import spaces
import numpy as np

class RecommenderSystemEnv(gym.Env):
  
    def __init__(self, n_products, n_features):
        super(RecommenderSystemEnv, self).__init__()

        # Environment parameters
        self.n_products = n_products
        self.n_features = n_features

        # Action and observation spaces
        self.action_space = spaces.Discrete(n_products)
        self.observation_space = spaces.Dict({
            'product_features': spaces.MultiDiscrete([n_features] * 4),
            'user_purchase_history': spaces.MultiBinary(n_products),
            'previous_recommendations': spaces.Discrete(n_products),
            'user_feedback': spaces.Discrete(2),  # Binary feedback (satisfied or dissatisfied)
            'time_step': spaces.Discrete(100)  # Assuming a maximum of 100 time steps
        })

        # Initialize environment state variables
        self.user_purchase_history = np.zeros(n_products, dtype=int)
        self.previous_recommendations = np.zeros(n_products, dtype=int)
        self.time_step = 0

    def _perform_transition(self, action):
        # Generate observation based on the action (product recommended)
        observation = {
            'product_features': np.random.randint(0, self.n_features, size=4),
            'user_purchase_history': self.user_purchase_history,
            'previous_recommendations': self.previous_recommendations,
            'user_feedback': np.random.choice([0, 1]),  # Simulating user feedback
            'time_step': self.time_step
        }

        # Update environment state variables
        self.previous_recommendations[action] = 1
        self.time_step += 1

        return observation

    def _get_reward(self, action, observation, terminated):
        # Calculate reward based on user feedback
        reward = observation['user_feedback'] if not terminated else 0
        return reward

    def _is_terminated(self, observation):
        # Terminate episode after a certain number of time steps (e.g., 100)
        return observation['time_step'] >= 99
        #return False

    def _get_info(self):
        # Additional information (if needed)
        return {}

    def step(self, action):
        observation = self._perform_transition(action)
        terminated = self._is_terminated(observation)
        reward = self._get_reward(action, observation, terminated)
        info = self._get_info()

        return observation, reward, terminated, info

    def reset(self, options=None):

        # Reset environment state variables
        self.user_purchase_history = np.zeros(self.n_products, dtype=int)
        self.previous_recommendations = np.zeros(self.n_products, dtype=int)
        self.time_step = 0

        # Initial observation
        observation = {
            'product_features': np.random.randint(0, self.n_features, size=4),
            'user_purchase_history': self.user_purchase_history,
            'previous_recommendations': self.previous_recommendations,
            'user_feedback': np.random.choice([0, 1]),
            'time_step': self.time_step
        }

        return observation, {}
\end{lstlisting}

\end{document}